\documentclass[letterpaper, 10 pt, conference]{ieeeconf}  % Comment this line out if you need a4paper

\IEEEoverridecommandlockouts                              % This command is only needed if 
\usepackage{amsmath}
\usepackage{mathtools}
\usepackage{tabularx}
\usepackage{amsmath, amssymb}%, amsthm}
\usepackage{hyperref, graphicx, cite, multirow}
\usepackage{subcaption}
\hypersetup{
	colorlinks=true,
	linkcolor=blue,
	citecolor=blue,
	urlcolor=blue,
}
\title{\LARGE \bf
AutoVRL: A High Fidelity Autonomous Ground Vehicle Simulator for Sim-to-Real Deep Reinforcement Learning }

\author{Shathushan Sivashangaran, Apoorva Khairnar and Azim Eskandarian
\thanks{\textit{Corresponding author: Shathushan Sivashangaran}}%
\thanks{The authors are with the Autonomous Systems and Intelligent Machines Laboratory, Virginia Tech, Blacksburg, VA 24061, USA.\indent(email: \href{mailto:shathushansiva@vt.edu}{shathushansiva@vt.edu}; \href{mailto:apoorvak@vt.edu}{apoorvak@vt.edu}; \href{mailto:eskandarian@vt.edu}{eskandarian@vt.edu}).} 
\thanks{AutoVRL source code and installation guide available at: \href{https://github.com/Shathushan-Sivashangaran/AutoVRL}{https://github.com/Shathushan-Sivashangaran/AutoVRL}}
\thanks{© 2023 the authors. This work has been accepted to IFAC for publication under a Creative Commons License CC-BY-NC-ND.}}

\begin{document}

\maketitle
\thispagestyle{empty}
\pagestyle{empty}

%%%%%%%%%%%%%%%%%%%%%%%%%%%%%%%%%%%%%%%%%%%%%%%%%%%%%%%%%%%%%%%%%%%%%%%%%%%%%%%%
\begin{abstract}

Deep Reinforcement Learning (DRL) enables cognitive Autonomous Ground Vehicle (AGV) navigation utilizing raw sensor data without a-priori maps or GPS, which is a necessity in hazardous, information poor environments such as regions where natural disasters occur, and extraterrestrial planets. The substantial training time required to learn an optimal DRL policy, which can be days or weeks for complex tasks, is a major hurdle to real-world implementation in AGV applications. Training entails repeated collisions with the surrounding environment over an extended time period, dependent on the complexity of the task, to reinforce positive exploratory, application specific behavior that is expensive, and time consuming in the real-world. Effectively bridging the simulation to real-world gap is a requisite for successful implementation of DRL in complex AGV applications, enabling learning of cost-effective policies. We present AutoVRL, an open-source high fidelity simulator built upon the Bullet physics engine utilizing OpenAI Gym and Stable Baselines3 in PyTorch to train AGV DRL agents for sim-to-real policy transfer. AutoVRL is equipped with sensor implementations of GPS, IMU, LiDAR and camera, actuators for AGV control, and realistic environments, with extensibility for new environments and AGV models. The simulator provides access to state-of-the-art DRL algorithms, utilizing a python interface for simple algorithm and environment customization, and simulation execution.

\end{abstract}

%%%%%%%%%%%%%%%%%%%%%%%%%%%%%%%%%%%%%%%%%%%%%%%%%%%%%%%%%%%%%%%%%%%%%%%%%%%%%%%%
\section{INTRODUCTION}

DRL is an effective technique for AGV navigation, enabling intelligent action decisions in a continuous and high-dimensional action space that comply with a learned policy, solely utilizing high-dimensional sensory input data. However, an optimal DRL policy is application specific, and can require days or weeks of continuous training \cite{sivashangaran2023deep, sivashangaran2021intelligent}.

Bridging the simulation to real-world gap is imperative for effective deployment of Deep Reinforcement Learning (DRL) for Autonomous Ground Vehicle (AGV), and other robotics applications \cite{zhao2020sim, muratore2019assessing, salvato2021crossing, ju2022transferring}. The substantial training time, and possible physical damage caused by collisions during training to reinforce application specific positive behavior renders real-world training markedly expensive. 

Cognitive navigation that utilizes raw sensor data with no dependency on a-priori maps or GPS is a requisite for AGV applications in information poor, dynamically altering environments such as rescue operations in regions where natural disasters occur, mapping unexplored subterranian environments, and extraterrestrial planetary exploration.

A diverse variety of proprietary and open-source simulation tools are available for rigid-body dynamics. For sim-to-real applications, high physical accuracy is required of the physics engine. Common physics engines and simulators utilized by researchers include Bullet, Gazebo, MuJoCo and Unreal Engine \cite{ferigo2020gym, li2022metadrive, hu2021sim}. In addition to a physics engine, a simulation platform must include high-fidelity 3D environment and AGV models to construct a digital twin.

We present AutoVRL (AUTOnomous ground Vehicle deep Reinforcement Learning simulator), an open-source high fidelity simulator for sim-to-real AGV DRL research and development. It is built upon the Bullet physics engine \cite{coumans2021} utilizing the OpenAI Gym \cite{openaigym} python library to provide an Application Programming Interface (API) for communication between DRL algorithms and environments. State-of-the-art DRL algorithms are implemented using Stable Baselines3 (SB3) \cite{stable-baselines3} which utilizes the PyTorch Machine Learning (ML) framework \cite{paszke2019pytorch} based on the Torch library. The source code and installation guide can be accessed using the link provided in the footnote. 

AutoVRL is equipped with sensor implementations of Global Positioning System (GPS), Inertial Measurement Unit (IMU), LiDAR and camera, high-fidelity model of the XTENTH-CAR AGV platform \cite{sivashangaran2022xtenth}, and realistic environments for training and evaluation. The simulator is extensible, with new environments and AGV models straightforward to add. Dependencies are quick to install, and the provided python interface enables simple SB3 algorithm customization, implementation and evaluation of different DRL libraries or novel DRL algorithms, and execution of simulations.  

The similar software frameworks in the simulator and real-world AGV, in Ubuntu with a python interface facilitate simulation to real-world DRL research. Policies learned in simulation are transferable to hardware using an identical software architecture, by swapping the PyBullet python bindings for the Bullet physics engine with a hardware interface, such as the commonly used Robot Operating System (ROS) framework \cite{quigley2009ros, macenski2022ros2}, for real-world action control inputs, and sensor observations. 

The rest of the paper is organized as follows: Section 2 details the simulator's architecture. Environment and AGV models are discussed in Section 3. The sensor implementations of GPS, IMU, LiDAR and camera are presented in Section 4. Section 5 describes actuation in AutoVRL. Training results in simulation and performance of simulator to hardware policy transfer are discussed in Section 6, and Section 7 concludes the paper.

\section{ARCHITECTURE}

\begin{figure}[!h]
    \centering
    \includegraphics[width = 0.7\columnwidth]{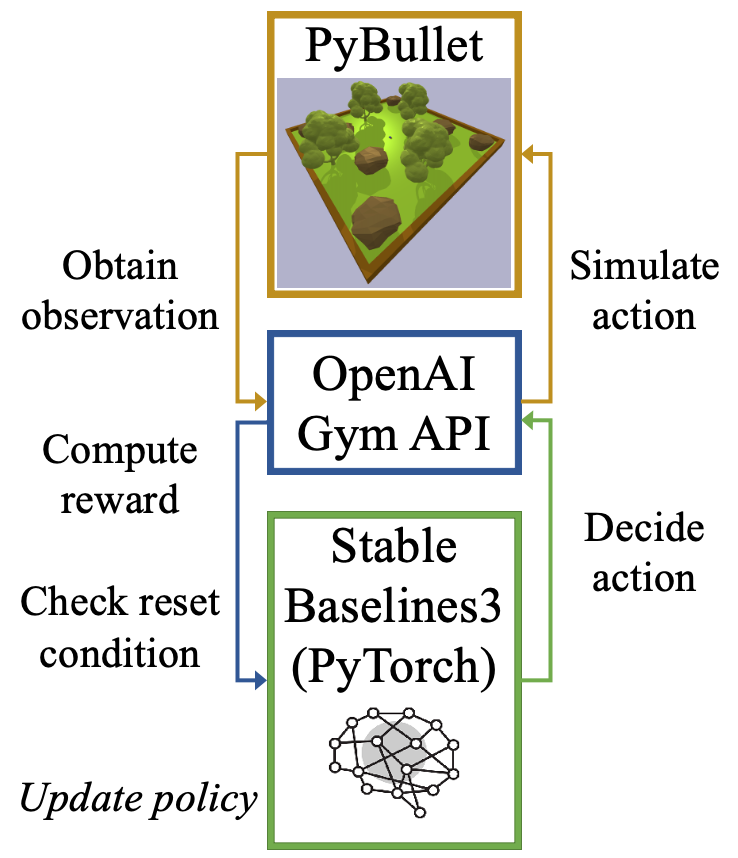}
    \caption{Schematic of AutoVRL architecture.} \label{Arch}
\end{figure}

AutoVRL comprises three components: PyBullet, OpenAI Gym and SB3. High-fidelity actions and observations are performed in the Bullet physics engine using PyBullet python bindings, and DRL is implemented via SB3, a set of customizable algorithms based on the PyTorch ML framework. The OpenAI Gym API communicates between the DRL algorithm and the Bullet physics engine. SB3 is easy to replace with alternate DRL libraries or custom DRL algorithms developed in PyTorch or TensorFlow for research evaluation. Figure \ref{Arch} illustrates AutoVRL's architecture and data flow between modules.

Goal specific actions dependent on the reward function are strategized by the DRL algorithm each time step, and simulated in the PyBullet simulation environment via Gym. The updated observations are relayed back from the physics engine to Gym using which the episode termination condition is verified within Gym. This information is transmitted to the SB3 DRL algorithm for learning and policy updates.

The policy learned in simulation is applied to a real-world AGV using the same architecture, by replacing the PyBullet functions for action control inputs and sensor observations with the AGV's hardware interface, such as the ROS framework. In previous research \cite{sivashangaran2021intelligent}, we utilized Simulink and the MATLAB Robotics System \cite{matlabRob} and Reinforcement Learning \cite{matlabRL} Toolboxes for DRL investigations, however, the use of MATLAB is computationally demanding and not optimal for real-time control of physical AGVs. AutoVRL operates on Ubuntu and provides a Python interface, commonly used in AGVs, enabling sim-to-real DRL research.

\section{AGV AND ENVIRONMENT MODELS}

High fidelity AGV and environment models are generated from 3D CAD models using Unified Robot Description Format (URDF) files, an eXtensible Markup Language (XML) file type that is also utilized in alternate simulation tools such as Gazebo. The URDF files for AGV and environment components include physical descriptions of each sub-component such as dimensions, joint type, color, and base position in the global coordinate system.

\begin{figure}[htbp]
\centering
\begin{minipage}{0.25\textwidth}
  \centering
\includegraphics[width=0.75\textwidth]{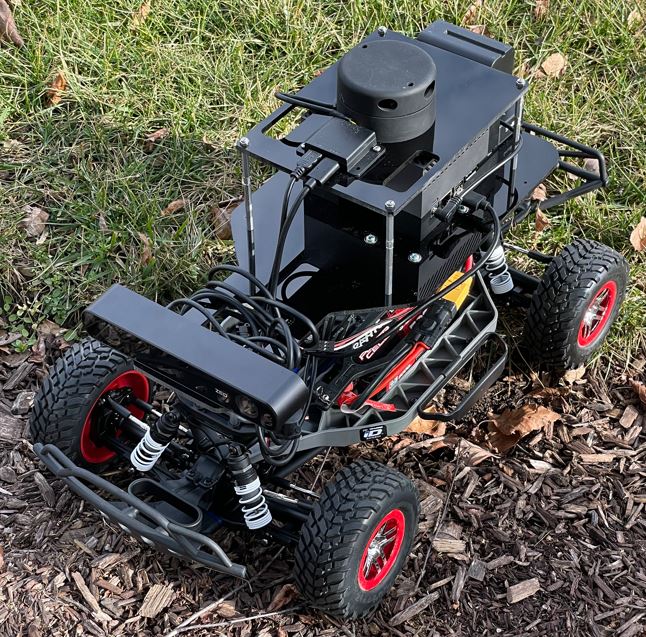}
\subcaption{}
\end{minipage}%
\begin{minipage}{0.25\textwidth}
  \centering
\includegraphics[width=0.75\textwidth]{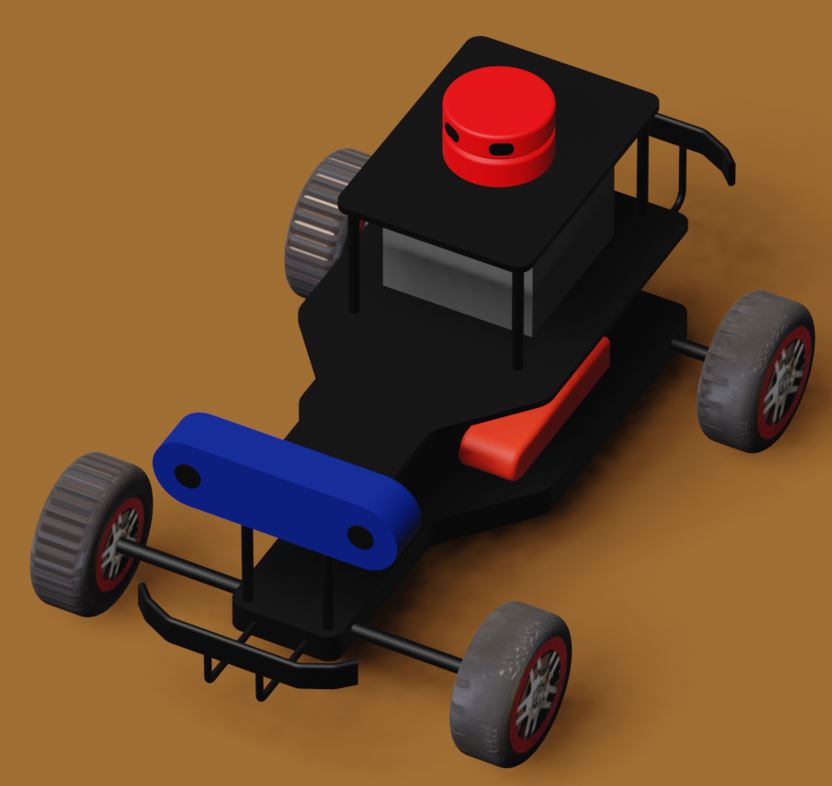}
\subcaption{}
\end{minipage}%
\caption{(a) XTENTH-CAR Ackermann steered AGV. (b) XTENTH-CAR digital twin. LiDAR marked in red, and camera in blue.} \label{xtenthcar}
\end{figure}

In order to facilitate simulation to real-world AGV research and development, AutoVRL includes a digital twin of XTENTH-CAR \cite{sivashangaran2022xtenth}, a proportionally 1/10th scaled Ackermann steered vehicle for connected autonomy and all terrain research developed with best-in-class embedded processing to facilitate real-world AGV DRL research. XTENTH-CAR shares similar hardware and software architectures to the full-sized X-CAR \cite{mehr2022XCAR} experimental vehicle, enabling DRL research for on-road Autonomous Vehicles (AVs) in addition to all-terrain AGVs. Figure \ref{xtenthcar} depicts the XTENTH-CAR AGV and digital twin. URDF files for digital twins of alternate, application specific AGVs are straightforward to generate, and load in AutoVRL using 3D CAD models. 

\begin{figure}[htbp]
\centering
\begin{minipage}{0.25\textwidth}
  \centering
\includegraphics[width=0.75\textwidth]{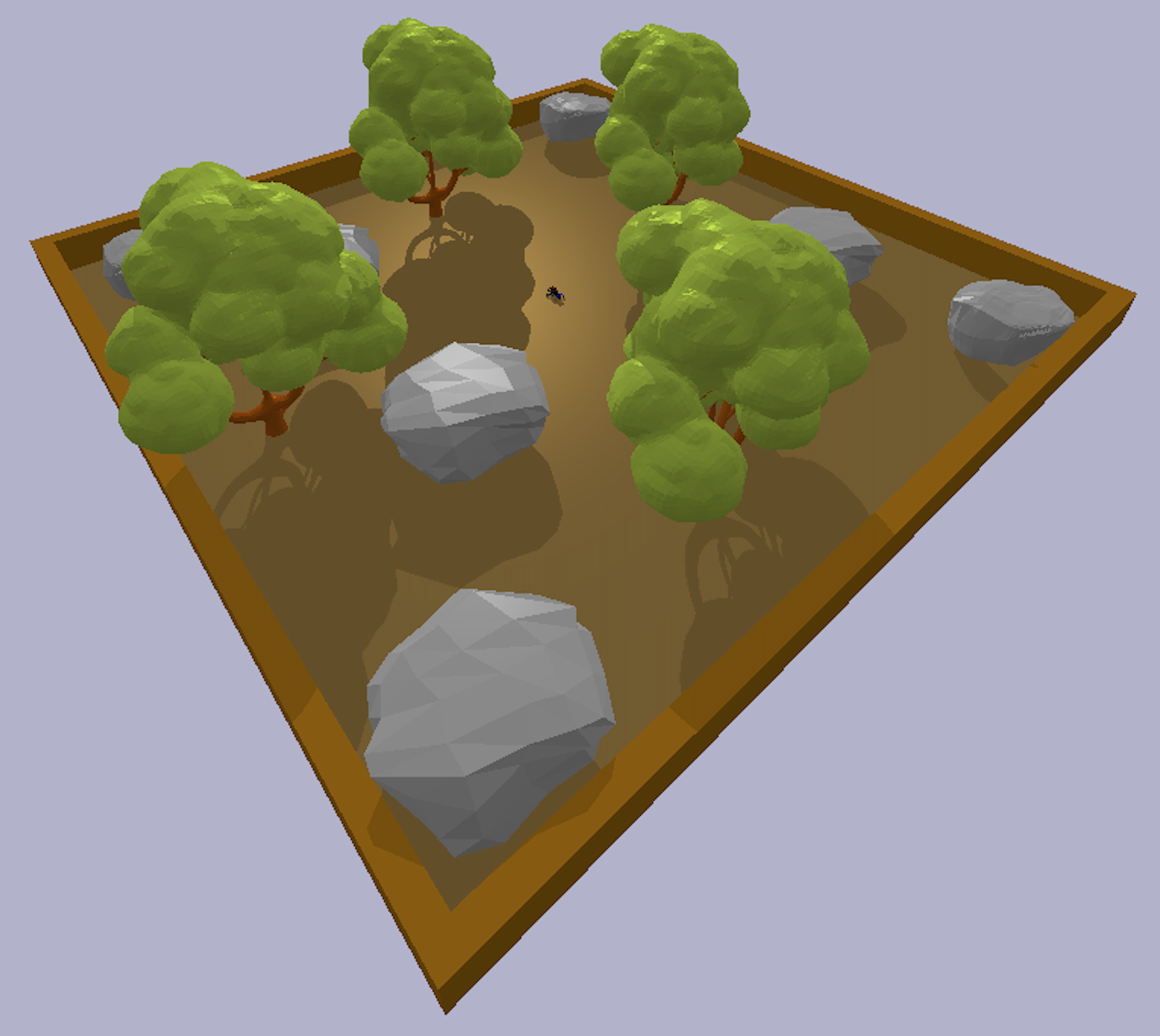}
\subcaption{}
\end{minipage}%
\begin{minipage}{0.25\textwidth}
  \centering
\includegraphics[width=0.7\textwidth]{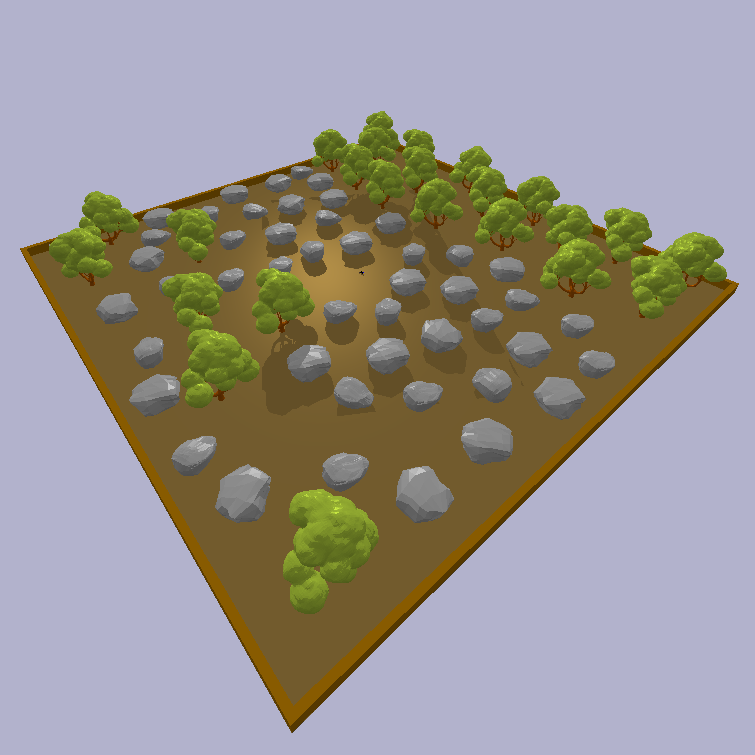}
\subcaption{}
\end{minipage}%
\caption{Outdoor environments with tree and boulder objects. (a) 20$m$ x 20$m$. (b) 50$m$ x 50$m$.} \label{envOut}
\end{figure}

\begin{figure}[htbp]
\centering
\begin{minipage}{0.25\textwidth}
  \centering
\includegraphics[width=0.7\textwidth]{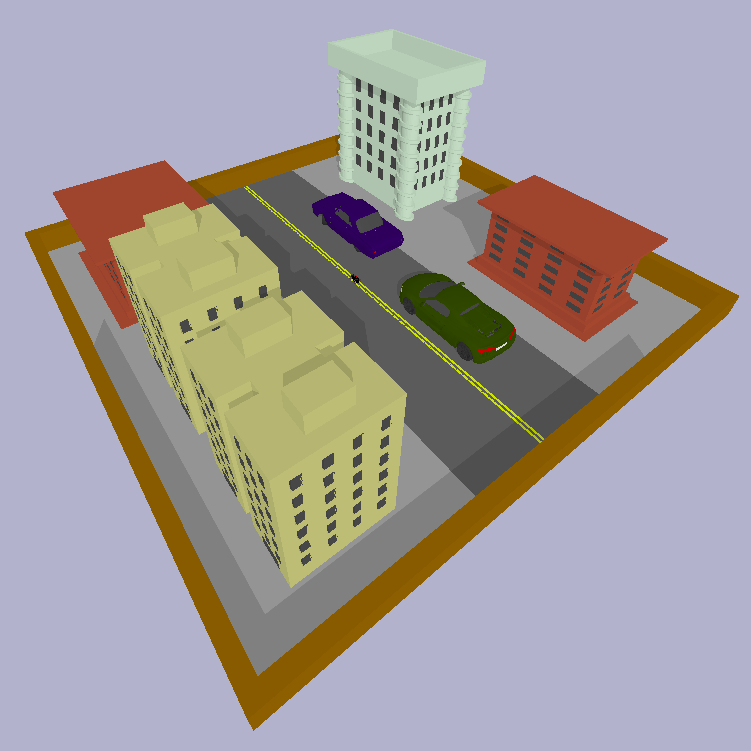}
\subcaption{}
\end{minipage}%
\begin{minipage}{0.25\textwidth}
  \centering
\includegraphics[width=0.7\textwidth]{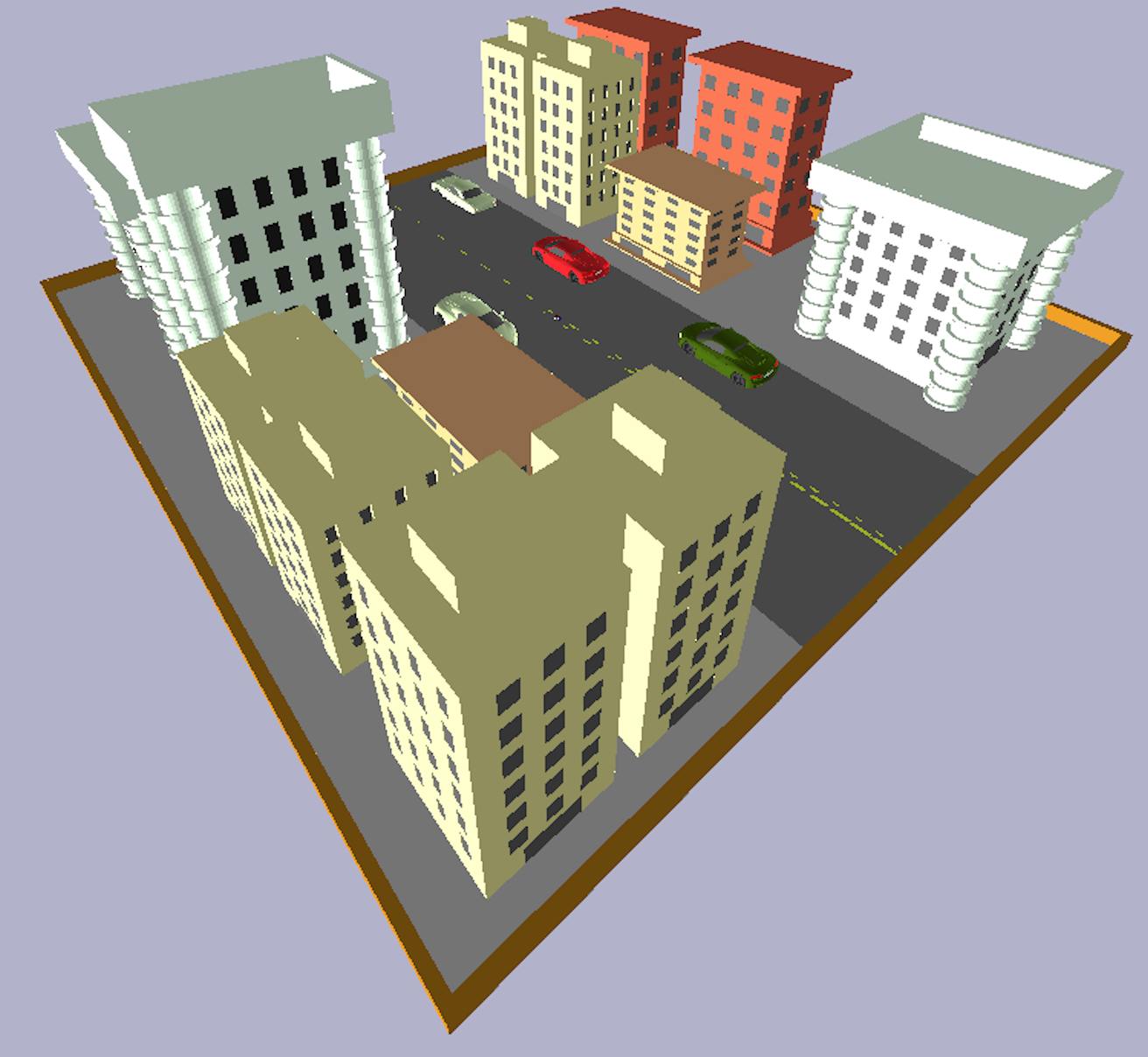}
\subcaption{}
\end{minipage}%
\caption{Urban environments with building and passenger vehicle objects. (a) 20$m$ x 20$m$. (b) 50$m$ x 50$m$.} \label{envUrb}
\end{figure}

\begin{figure}[!h]
    \centering
    \includegraphics[width = 0.4\columnwidth]{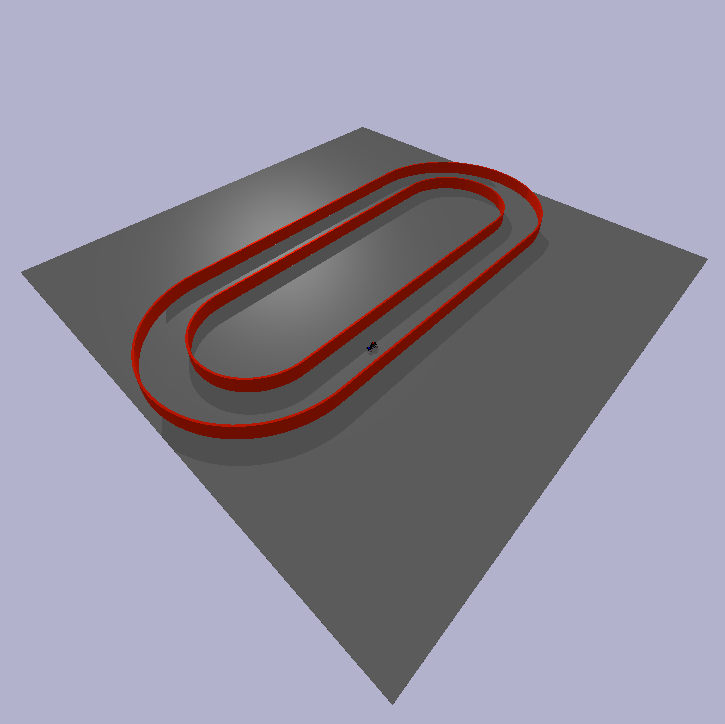}
    \caption{Oval race track environment.} \label{envRaceOval}
\end{figure}

Five environments are provided for training and evaluation. These include 20$m$ x 20$m$ and 50$m$ x 50$m$ outdoor and urban environments with realistic objects such as trees and boulders in the outdoor map, and buildings and passenger vehicles in the urban scenario, and an oval race track. The environments are illustrated in Figures \ref{envOut} - \ref{envRaceOval}.

The environments are simple to enlarge for larger AGVs, by modifying the URDF files, and adding additional objects to populate the map. Application specific scenarios, such as indoor household or office, and subterranean environments can be generated with open-source, or custom CAD models.

\section{SENSING}

This section presents AutoVRL's sensor suite. The simulator is equipped with implementations of key AGV sensors that include GPS, IMU, LiDAR and camera.

\subsection{GPS}

GPS is a Global Navigation Satellite System (GNSS) that provides precise geo-spatial positioning. Global position and orientation of the simulated AGV are extracted using the PyBullet getBasePositionAndOrientation function. The position $(x, y, z)$ is in Cartesian world coordinates, and the orientation is a quaternion of the form $(x, y, z, w)$. The getEulerFromQuaternion PyBullet function is used to convert the quaternion orientation to Euler angles, roll, pitch and yaw $(\phi, \theta, \psi)$.

\subsection{IMU}

IMUs measure acceleration, angular rates and orientation. Linear and angular velocities in Cartesian global coordinates of the form $(\dot{x}, \dot{y}, \dot{z})$ and $(\dot{\phi}, \dot{\theta}, \dot{\psi})$ are acquired using the getBaseVelocity PyBullet function. Accelerations, both linear and angular, are computed using this information, and the sampling time $t_{s}$.

\subsection{LiDAR}

\begin{figure}[!h]
    \centering
    \includegraphics[width = 0.4\columnwidth]{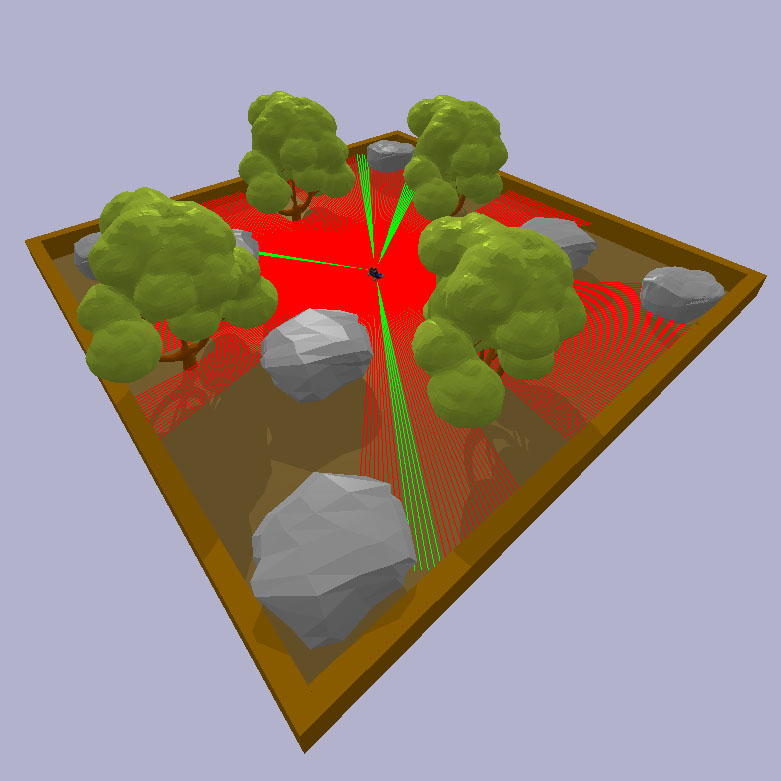}
    \caption{Ray tracing implementation of LiDAR. Red rays hit obstacles, and green rays detect none.} \label{lidar}
\end{figure}

LiDAR (Light Detection and Ranging) detects and determines distances of objects from the ego vehicle by measuring the time taken for reflected pulsed laser beams to return to the system \cite{raj2020survey}. AutoVRL utilizes ray tracing, a technique to model light transport, and compute light rays hitting and reflecting off surfaces \cite{wald2009state}, to simulate LiDAR. The ray tracing implementation of LiDAR in AutoVRL is illustrated in Figure \ref{lidar} where the red rays convey detected objects, and green rays miss obstacles.

The origin coordinates of the rays $(x_{RF}, y_{RF}, z_{RF})$ are computed as follows,

\begin{equation} \label{}
x_{RF} = x_{L,A}cos(\psi_{A}) + x_{A}
\end{equation}

\begin{equation} \label{}
y_{RF} = y_{L,A}sin(\psi_{A}) + y_{A}
\end{equation}

\begin{equation} \label{}
z_{RF} = z_{L,A} + z_{A}
\end{equation}

Here $x_{L,A}$, $y_{L,A}$ and $z_{L,A}$ are coordinates of the LiDAR with respect to the AGV's body frame, $x_{A}$, $y_{A}$ and $z_{A}$ represent the position of the AGV in the global coordinate frame, and $\psi_{A}$ is the AGV's yaw.

The position of the end point of $n$ rays $(x_{RT}, y_{RT}, z_{RT})$ with a maximum ray length $r_{max}$ are generated as follows,

\begin{equation} \label{}
x_{RT} = r_{max}sin\left((\frac{\pi}{4} - \psi_{A}) + \alpha(\frac{1:n}{n})\right)
\end{equation}

\begin{equation} \label{}
y_{RT} = r_{max}cos\left((\frac{\pi}{4} - \psi_{A}) + \alpha(\frac{1:n}{n})\right)
\end{equation}

\begin{equation} \label{}
z_{RT} = z_{L,A}
\end{equation}

Here $(\frac{\pi}{4} - \psi_{A})$ represents the LiDAR's orientation, and $\alpha(\frac{1:n}{n})$ segments a specified angle $\alpha$, which in Figure \ref{lidar} is $\frac{\pi}{2}$, to $n$ equally spaced rays.

The hit position of each ray $(x_{HP}, y_{HP})$ is obtained from the PyBullet rayTestBatch function, using which the euclidean distance to detected obstacles $r_{o}$ is determined,

\begin{equation} \label{}
r_o = \sqrt{(x_{HP} - x_{RF})^{2} + (y_{HP} - y_{RF})^{2}}
\end{equation}

\subsection{Camera}

Cameras convert light captured from the surrounding environment into electric signals for image processing. Images are used to detect and identify objects, track dynamic objects, and estimate pose via visual odometry.

AutoVRL renders images from an arbitrary AGV camera position using PyBullet's OpenGL GPU visualizer and CPU renderer. Two 4 by 4 matrices, the view matrix and the projection matrix are used to specify the synthetic camera using the computeViewMatrix and computeProjectionMatrixFOV PyBullet functions.

%The view matrix is computed via the PyBullet computeViewMatrix function using the eye position, and focus point of the camera in Cartesian world coordinates. PyBullet's computeProjectionMatrixFOV function is used to determine the projection matrix utilizing camera specifications such as Field of View (FoV) and aspect ratio.

Images are generated from the view and projection matrices utilizing the getCameraImage PyBullet function, which returns a RGB image, a depth buffer, and a segmentation mask buffer with unique object IDs of detected objects at each pixel. %Figures \ref{camOutdoor} and \ref{camUrban} depict synthetic camera images from the ZED 2 camera of XTENTH-CAR's digital twin in outdoor and urban environments. 

\section{ACTUATION}

The action outputs from the DRL algorithm are normalized in the range [-1 1]. These are converted to vehicle control inputs, which for an Ackermann steered AGV are throttle $T$ in the range [0 1], and steering angle $\delta$ in the range $[\delta_{min}  \delta_{max}]$ computed as follows,

\begin{equation} \label{}
T = min(max(a_{T}, 0), 1)
\end{equation}

\begin{equation} \label{}
\delta = max(min(a_{\delta}, \delta_{max}), \delta_{min})
\end{equation}

Here $a_{T}$ and $a_{\delta}$ are the normalized action outputs strategized by the DRL policy.

The vehicle control inputs are applied to the simulated AGV using the setJointMotorControlArray PyBullet function. The number of joints the control inputs are applied to is dependent on the AGV model. For an Ackermann steered AGV, the steering angle is applied to two joints, and joint speed to four. %Figure \ref{actuation} provides a generalized schematic of actuation in AutoVRL agnostic of AGV model. 

%\begin{figure}[!h]
%    \centering
%    \includegraphics[width = 0.9\columnwidth]{Images/actuation.png}
    %\setlength{\abovecaptionskip}{-\baselineskip}
%    \caption{Actuation in AutoVRL agnostic of AGV model.} \label{actuation}
%\end{figure}

Resistive forces are accounted for to improve the fidelity of the simulated action. Friction $f$ is computed for the current time step using drag constants $C_d$ and $C_r$, and the joint speed from the previous time step $v_{j,i-1}$,

\begin{equation} \label{}
f = v_{j,i-1}(v_{j,i-1}C_d + C_r)
\end{equation}

The acceleration of each joint is determined as follows, from the throttle input $T$, throttle constant $C_T$ and resistive force $f$,

\begin{equation} \label{}
\dot{v}_{j,i} = C_{T}T - f
\end{equation}

The speed of each joint for the current time step $v_{j,i}$ is computed as,

\begin{equation} \label{}
v_{j,i} = v_{j,i-1} + t_{s}\dot{v_{j,i}}
\end{equation}

\section{RESULTS AND DISCUSSION}

In this section, we present training and evaluation results for two applications which each utilize a custom reward formulation. The first is a search application where the AGV is trained to explore shaded regions unobservable to aerial surveillance to detect subjects of interest. The exploration policy is learned using LiDAR range sensor observations, and Computer Vision (CV) is utilized to detect and identify subjects. The second policy is trained for autonomous racing. Moreover, the policies learned in simulation are implemented on a physical XTENTH-CAR to gauge real-time computation performance. An Intel Core i9 13900KF CPU and NVIDIA GeForce RTX 4090 GPU were used for training.

\subsection{Exploration and Search}

AGVs frequently work in tandem with Unmanned Aerial Vehicles (UAVs) in search and exploration applications such as military and Search and Rescue (SAR) operations. AGVs are especially beneficial in exploring regions with poor aerial visibility such as areas with thick foliage, and high obstacle density. The reward $R_{search}$ was formulated to promote exploration in dense, obstacle filled terrain as follows, and trained using the Soft Actor-Critic (SAC) \cite{haarnoja2018soft} algorithm which we found to be the best sample efficient off-policy DRL algorithm for AGV navigation in prior work \cite{sivashangaran2023deep}. 

\begin{equation} \label{}
    R_{search} = 
\begin{dcases}
    0.005r_{m}^{2} + 5.0T^{2} - 2.0\delta^{2} &\\ 
    2.0              & \text{if } 2.0 < r_{m} < 2.5\\
    -50              & \text{if } r_{m} < 1.0\\
\end{dcases}
\end{equation}

Here $r_{m}$ is the minimum LiDAR range distance, which determines the AGV's displacement to the nearest obstacle. A positive reward of 0.005 was applied to the square of $r_{m}$ to incentivize the agent to traverse trajectories devoid of obstacles. Positive and negative rewards of 5.0 and -2.0 were applied to throttle, $T$ and steering angle, $\delta$ control inputs respectively to encourage exploration, and prevent learning of a sub-optimal trajectory with repeated circular motion in the same vicinity. A reward value of 2.0 was assigned when the agent was between 2.0 and 2.5 $m$ of the nearest object to promote exploration near obstacles, that are often inaccessible to aerial surveillance. A penalty of 50.0 was applied when the agent was within 1 $m$ of the nearest obstacle to improve the safety of exploration trajectories.

\begin{figure}[!h]
    \centering
    \includegraphics[width = 0.85\columnwidth]{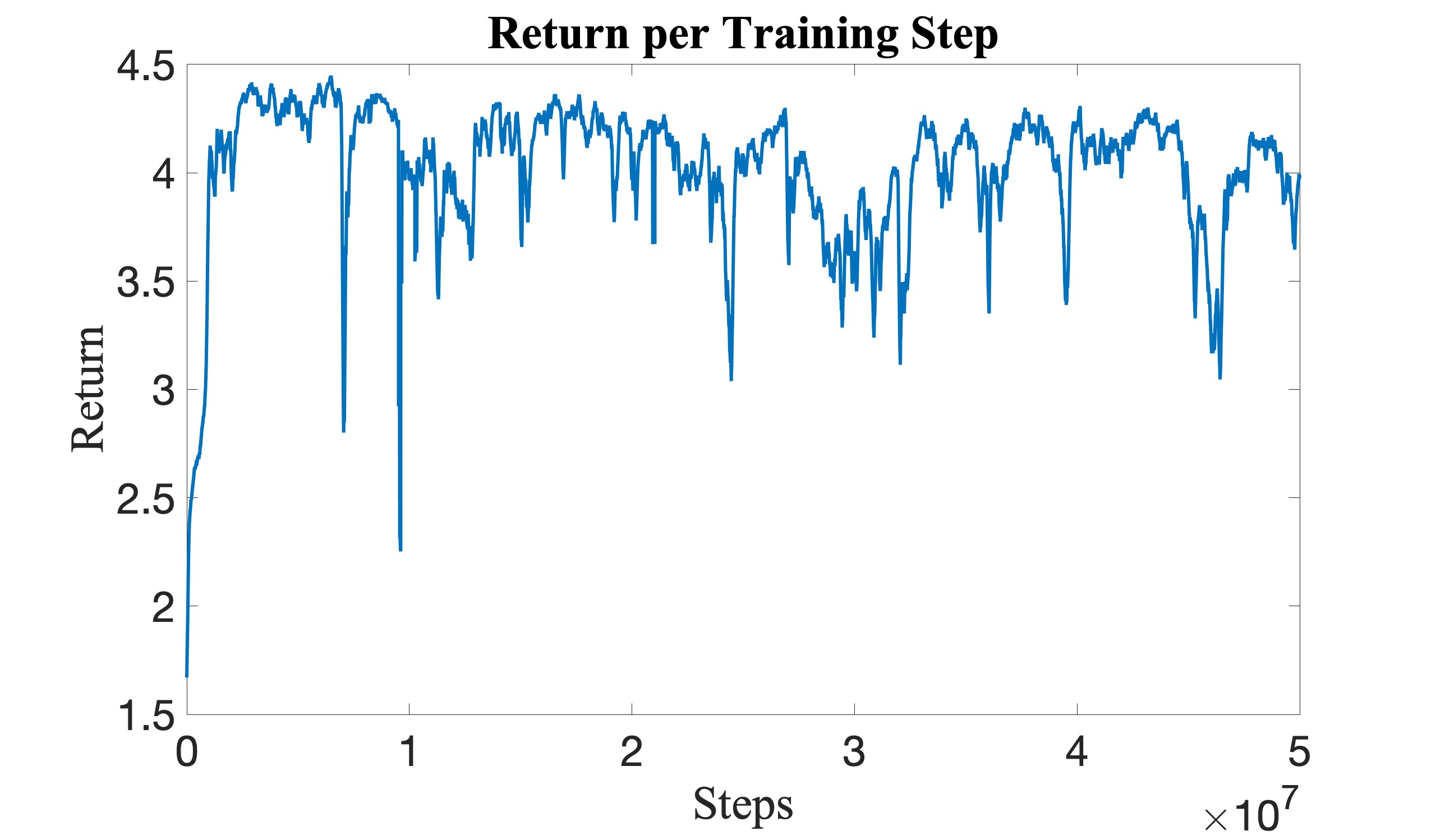}
    \caption{Order 100,000 moving average return during training in the 20$m$ x 20$m$ outdoor environment.} \label{return_plot}
\end{figure}

The default SB3 SAC hyperparameters were used to train the agent for 50,000,000 steps over 9.5 days in the 20$m$ x 20$m$ outdoor environment. The learning curve is shown in Figure \ref{return_plot}.

\begin{figure}[!h]
    \centering
    \includegraphics[width = 0.58\columnwidth]{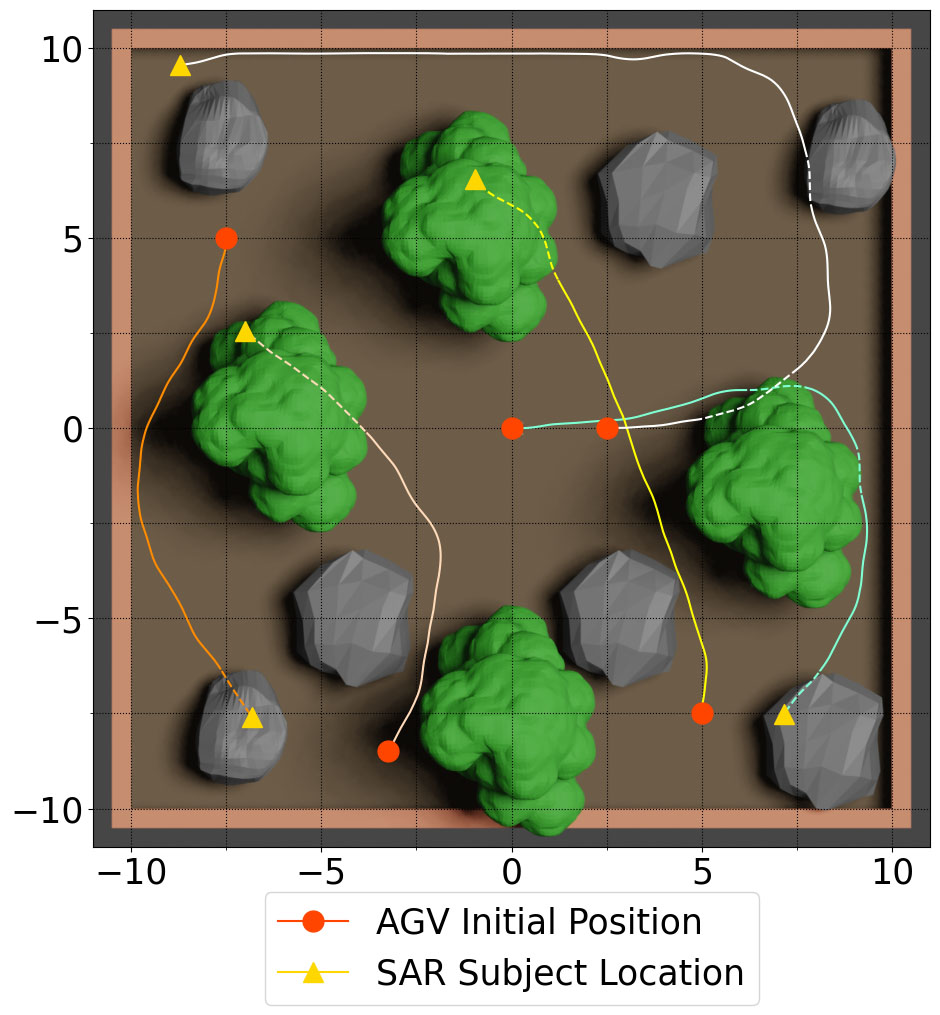}
    \caption{Post-training AGV trajectories in the outdoor environment from various initial positions.} \label{outenv_traj}
\end{figure}

The order 100,000 moving average return increased rapidly during the first 1,000,000 training steps, and fluctuated around a value of 3.9 over the remainder of the training period. The minimum average return increases as training progresses which indicates that an improved policy with better convergence can be learned over a longer training period. Figure \ref{outenv_traj} illustrates single vehicle post-training trajectories in the training environment evaluated at various combinations of AGV initial position and SAR subject location.

\begin{figure}[!h]
    \centering
    \includegraphics[width = 0.58\columnwidth]{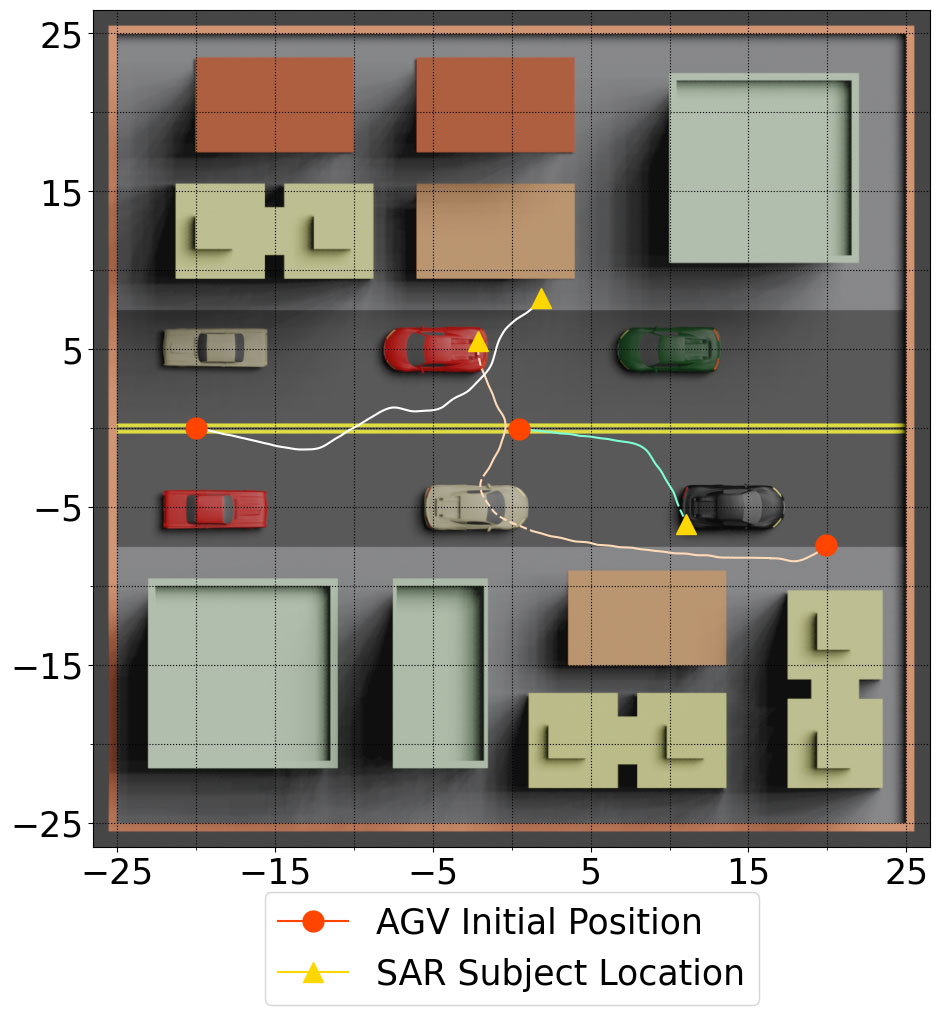}
    \caption{Post-training AGV trajectories in the urban environment from various initial positions.} \label{urbenv_traj}
\end{figure}

The AGV locates SAR subjects underneath trees and boulders that are unobservable to aerial surveillance, while maintaining a safe distance to obstacles solely utilizing raw sensor data with no prior knowledge of map characteristics. The trained agent was further evaluated in the 50$m$ x 50$m$ urban environment to test the extensibility, and robustness of the learned policy. Figure \ref{urbenv_traj} illustrates single vehicle post-training trajectories in the evaluation environment.

The difference in map and object characteristics in the evaluation environment presented a new challenge, however, the agent was successfully able to strategize efficient, collision free trajectories to explore shaded regions beneath buildings, and underneath vehicles. A longer training period, tuned hyperparameters and refined reward function will enable more efficient learning, and yield an improved policy to chart more complex trajectories to explore the surrounding environment indefinitely.

\subsection{Autonomous Racing}

\begin{figure}[!h]
    \centering
    \includegraphics[width = 0.85\columnwidth]{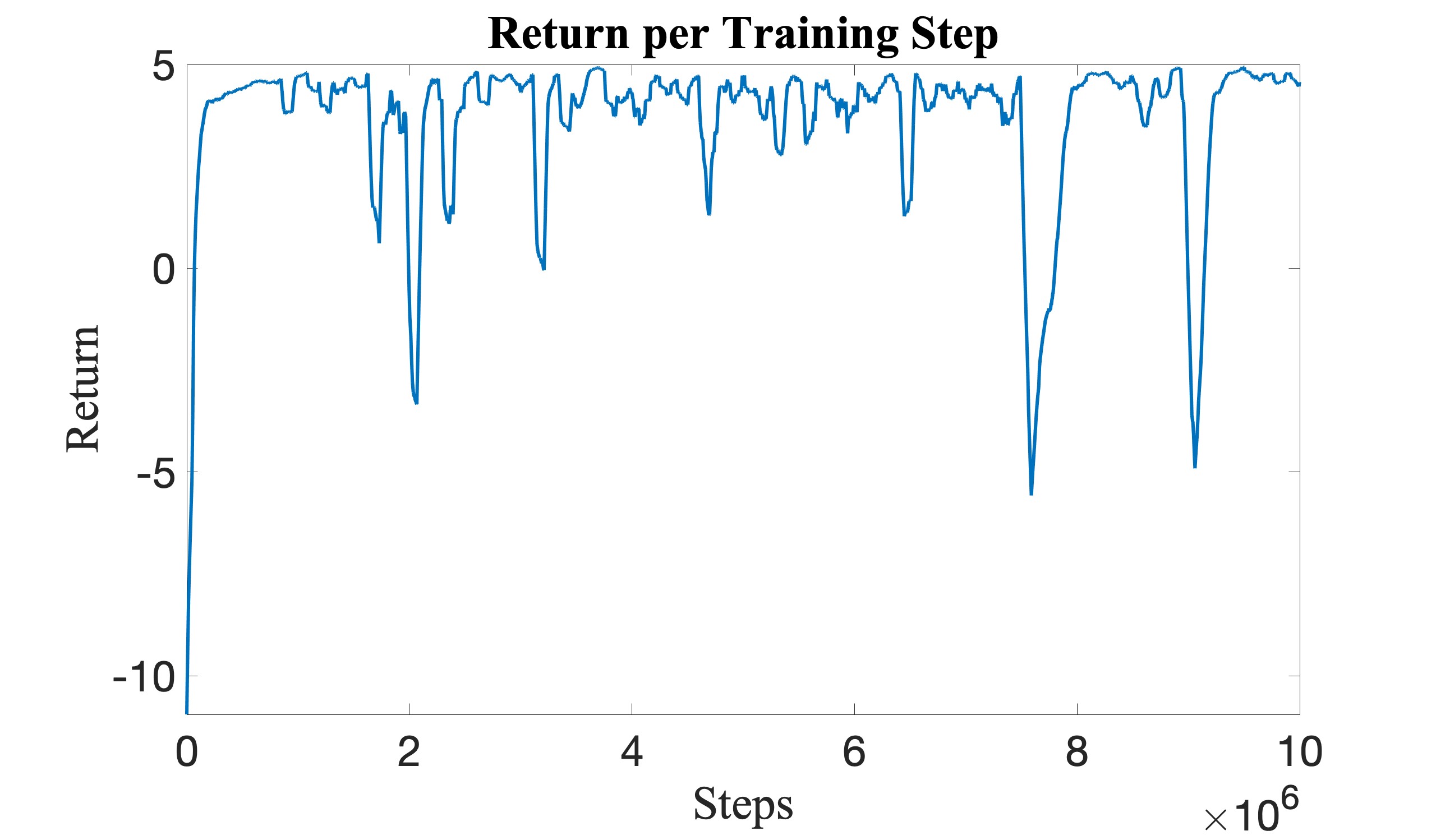}
    \caption{Order 100,000 moving average return during training in the oval racetrack.} \label{return_plot_race}
\end{figure}

\begin{figure}[!h]
    \centering
    \includegraphics[width = 0.58\columnwidth]{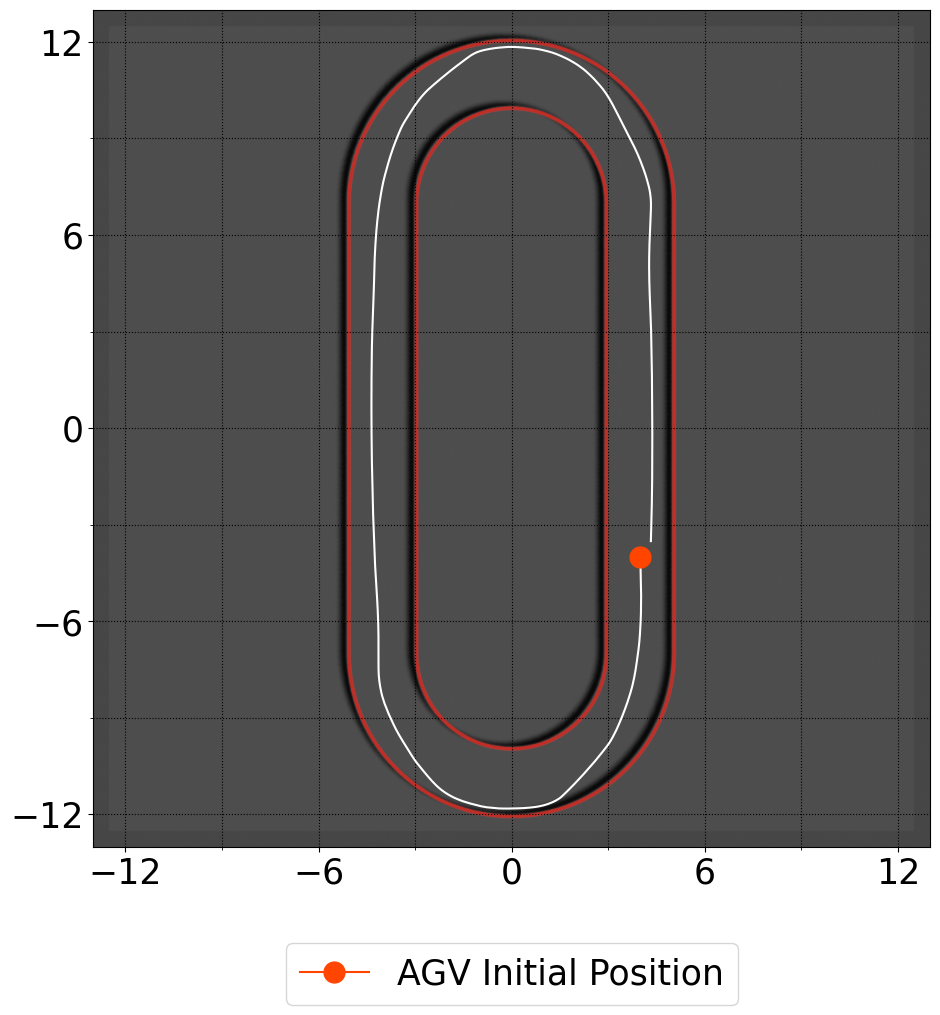}
    \caption{Post-training AGV trajectory in the oval racetrack.} \label{raceenv_traj}
\end{figure}

Motorsport drives automobile innovation, and autonomous racing has gained increasing popularity in recent years. A conservative reward function, $R_{racing}$ was formulated as follows to promote safe, time-optimal trajectories to gauge the effectiveness of the AutoVRL simulator. The agent was trained for 10,000,000 steps over 1.4 days using Proximal Policy Optimization (PPO) \cite{schulman2017proximal}, an on-policy DRL algorithm which is less sample efficient than SAC, but potentially more robust to hyperparameter tuning, which is of benefit to simulation to real-world applications. The learning curve and post-training trajectory are illustrated in Figures \ref{return_plot_race} and \ref{raceenv_traj}.

\begin{equation} \label{}
    R_{racing} = 
\begin{dcases}
    5.0T^{2} - 2.0\delta^{2} &\\ 
    -50              & \text{if } r_{m} < 0.8\\
\end{dcases}
\end{equation}

The order 100,000 moving average return rapidly increased during the first 600,000 training steps to a value of 4.6. The trajectory at the end of training was collision-free, but not time-optimal as the agent traversed around the longer outside curve of the corner. A time-optimal trajectory in which the agent switches lanes to the outside, and swoops to the inside of the corner before exiting can be learned with a refined reward function. The results demonstrate the versatility of AutoVRL for a variety of AGV applications.

\subsection{Real-World Performance}

In order to test simulation to real-world performance, we implemented the policies learned in simulation on a physical XTENTH-CAR equipped with the NVIDIA Jetson AGX Orin System on Module (SOM). The AutoVRL ROS driver utilizes real-world sensor data to select actions using the policy learned in simulation, and applies the chosen control signals to the AGV's actuators in real-time. Table \ref{computation} lists the average CPU and memory usages for real-time actuator control, LiDAR Point Cloud Map (PCM), CV using Mask R-CNN and DRL for navigation.

\begin{table}[ht]
\centering
\caption{Real-World CPU and Memory Usage.}
\begin{tabular}[t]{ccc}
\hline
\textbf{Task} & \textbf{CPU Usage \%} & \textbf{Memory Usage \%} \\
\hline
Actuator Control, \\ PCM \& DRL & 7.6 & 12.6 \\
\hline
Actuator Control, \\ PCM, CV \& DRL & 24.2 & 17.0 \\
\hline
\end{tabular}
\label{computation}
\end{table}

Despite the high training cost, real-time control using a trained DRL policy is significantly less computationally demanding. The Jetson AGX Orin utilizes peak CPU and memory usage percentages of 24.2 and 17.0. The trained policy is under 10 MB, and works seamlessly in both the physics engine in which it can be trained for days or weeks, and on real-world hardware.
 
\section{CONCLUSIONS}

This paper introduced AutoVRL, a high-fidelity AGV simulator for simulation to real-world DRL research and development that is beneficial for DRL applications that require days or weeks of continuous training. The proposed simulator is built using open-source tools, and provides access to state-of-the-art DRL algorithms for a range of AGV applications. A digital twin of the XTENTH-CAR AGV and five 3D environments that include an oval racetrack, and outdoor and urban surroundings come with AutoVRL. The simulator is extensible, with application specific AGV and environment models straightforward to add. DRL agents for search applications, and autonomous racing were trained in AutoVRL using off-policy and on-policy algorithms, and shown to learn favorable policies using custom reward formulations. Moreover, the trained policies were implemented on the real-world XTENTH-CAR AGV to assess real-time performance on an embedded computer, and shown to utilize a fraction of the available CPU and memory resources. For future work, additional AGV and environment models will be added to AutoVRL.

\addtolength{\textheight}{-12cm}   % This command serves to balance the column lengths
                                  % on the last page of the document manually. It shortens
                                  % the textheight of the last page by a suitable amount.
                                  % This command does not take effect until the next page
                                  % so it should come on the page before the last. Make

%%%%%%%%%%%%%%%%%%%%%%%%%%%%%%%%%%%%%%%%%%%%%%%%%%%%%%%%%%%%%%%%%%%%%%%%%%%%%%%%

%%%%%%%%%%%%%%%%%%%%%%%%%%%%%%%%%%%%%%%%%%%%%%%%%%%%%%%%%%%%%%%%%%%%%%%%%%%%%%%%

%%%%%%%%%%%%%%%%%%%%%%%%%%%%%%%%%%%%%%%%%%%%%%%%%%%%%%%%%%%%%%%%%%%%%%%%%%%%%%%%
%\section*{APPENDIX}

%\section*{ACKNOWLEDGMENT}

%%%%%%%%%%%%%%%%%%%%%%%%%%%%%%%%%%%%%%%%%%%%%%%%%%%%%%%%%%%%%%%%%%%%%%%%%%%%%%%%

\bibliography{root}
\bibliographystyle{IEEEtran}

%\begin{thebibliography}{99}

%\end{thebibliography}

\end{document}